\documentclass{article}
\usepackage{ijcai22}

\usepackage{times}
\usepackage{soul}
\usepackage{url}
\usepackage[hidelinks]{hyperref}
\usepackage[utf8]{inputenc}
\usepackage[small]{caption}
\usepackage{graphicx}
\usepackage{amsmath}
\usepackage{amsthm}
\usepackage{booktabs}
\usepackage{algorithm}
\usepackage{algorithmic}
\usepackage{float}
\usepackage{multirow}
\usepackage{graphicx}
\usepackage{lscape}
\usepackage{amssymb}
\usepackage{multirow}
\usepackage{color,soul}
\usepackage{amsmath}
\usepackage{booktabs}
\usepackage{multirow}
\usepackage{bbding}
\usepackage{booktabs}
\usepackage{makecell}
\usepackage{color}

\title{Astock: A New Dataset and Automated Stock Trading based on Stock-specific News Analyzing Model}

\author{
Jinan Zou$^1$ \footnote{Both authors contributed equally}\and
Haiyao Cao$^1$ \footnotemark[1]\and
Lingqiao Liu$^1$\and
Yuhao Lin$^1$ \and
Ehsan Abbasnejad$^1$ \and
Javen Qinfeng Shi$^1$ \footnote{Corresponding Author} \and
\\
\affiliations
$^1$Australian Institute for Machine Learning, University of Adelaide\\
\emails
\{jinan.zou, haiyao.cao, lingqiao.liu, yuhao.lin01, ehsan.abbasnejad, javen.shi\}@adelaide.edu.au
}

\begin{document}

\maketitle

\begin{abstract}
Natural Language Processing(NLP) demonstrates a great potential to support financial decision-making by analyzing the text from social media or news outlets. In this work, we build a platform to study the NLP-aided stock auto-trading algorithms systematically. In contrast to the previous work, our platform is characterized by three features: (1) We provide financial news for each specific stock. (2) We provide various stock factors for each stock. (3) We evaluate performance from more financial-relevant metrics. Such a design allows us to develop and evaluate NLP-aided stock auto-trading algorithms in a more realistic setting. In addition to designing an evaluation platform and dataset collection, we also made a technical contribution by proposing a system to automatically learn a good feature representation from various input information. The key to our algorithm is a method called semantic role labeling Pooling (SRLP), which leverages Semantic Role Labeling (SRL) to create a compact representation of each news paragraph. Based on SRLP, we further incorporate other stock factors to make the final prediction. In addition, we propose a self-supervised learning strategy based on SRLP to enhance the out-of-distribution generalization performance of our system. Through our experimental study,  we show that the proposed method achieves better performance and outperforms all the baselines' annualized rate of return as well as the maximum drawdown of the CSI300 index and XIN9 index on real trading. Our Astock dataset and code are available at https://github.com/JinanZou/Astock.

\end{abstract}

\section{Introduction}
\begin{figure}[t]
\begin{center}  
   \includegraphics[width=\linewidth]{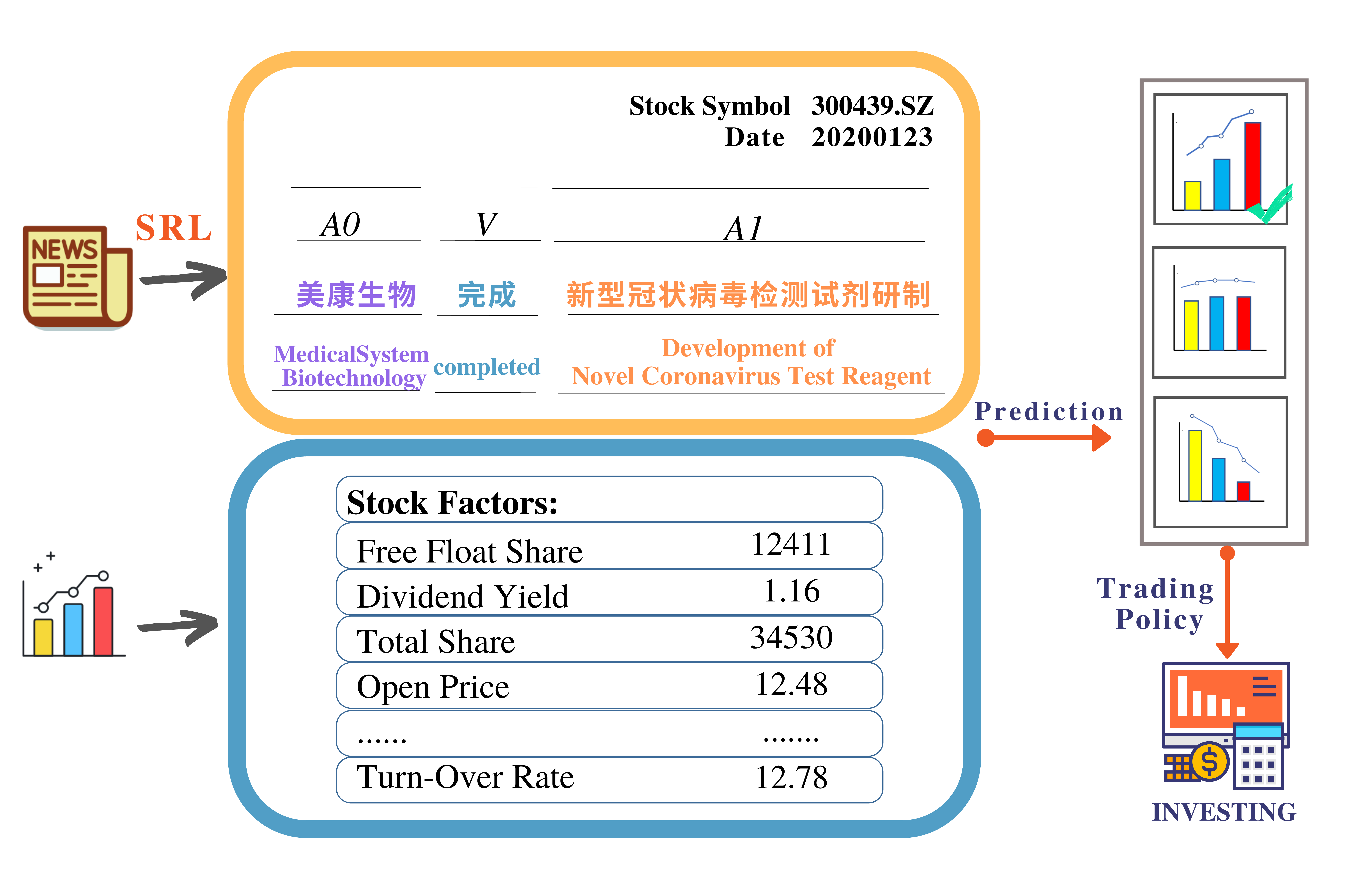}
\end{center}
\caption{Overview of the automated stock trading system.}
\label{tradingsystem}
\end{figure}
The Stock prediction has been an attractive task for a long time, and it is still challenging since the stochasticity of the market and behavior patterns of participators are fluctuating and elusive. Stock forecasting based on Natural Language Processing (NLP) techniques is a promising solution since text information, e.g., tweets, financial news etc., is strongly correlated with the stock prices. However, the NLP-based stock forecasting research is still scattered without unified definitions, benchmark datasets, clear articulations of the tasks, which severaly hinders progress of this field.

Existing approaches are usually based on market sentiment analysis \cite{xu-cohen-2018-stock,cheng2021modeling} and use news to predict the related securities' price on the following trading day(s) \cite{zhang2017stock,li2020role}. Despite the limited success in those studies, the existing works are still far from realistic for two reasons: Firstly, previous methods ignore the financial factors, which plays a key role in practical trading. Secondly, these models are evaluated only on intermediate performance metric, e.g., stock movement prediction accuracy. It is unclear how well they can support a practical trading system to make sufficient profit.

To address the problems above, we construct a China A-shares market dataset with news and stock factors called Astock. Specifically, we annotate all occurrences of the three trading actions (long, preserve, short) in 40,963 news originated from Tushare \footnote{http://tushare.org} with a valid official license, which describes the major financial events. The dataset also includes various stock factors to build a realistic system. Based on Astock, we establish a semantic role labeling pooling (SRLP) to build a compact representation for stock-specific news and predict the stock movement. This work also explores how to leverage a self-supervised method better to upgrade the SRLP method, which achieves better performance for classification and high domain generalization ability.

In experiments, we further propose a realistic trading platform that outperforms the state-of-the-art text classification baseline's average returns and Sharpe Ratios over the CSI300 index and XIN9 index of testing period from January 2021 to November 2021. Specifically, we analyze the profitability of the proposed strategy based on stock movement prediction result for real trading as shown in Figure \ref{tradingsystem}. The primary contributions of this work can be summarized as follows:
\begin{itemize}

\item We construct a brand new Chinese stock prediction task dataset with stock-specific news and stock factors.
\item Our SRLP characterizes the key attributes of financial events, which is convenient for incorporating other stock factors and further creating a self-supervised module on top of the SRLP method. Our self-supervised SRLP method obtains competitive stock movement prediction and out-of-distribution (OOD) generalization results.
\item We further evaluate algorithm performance on real-world trading from more financial-relevant metrics. By conducting extensive experimental studies, we show that our self-supervised SRLP achieves remarkable performance on these metrics. Furthermore, we observe that the proposed trading strategies work well in practice.
\end{itemize}

\section{Related Work}

\subsection{Text-based Stock Prediction}
In recent years, the use of text-based information, especially news and social media, has significantly improved the performance of stock prediction tasks and these methods usually rely on text-based features and sentiment analysis to forecast stock movements \cite{xu-cohen-2018-stock,hu2018listening,ding2015deep}. However, These approaches assume that the real-trading distribution was the same as the training distribution, which is not realistic as it is difficult to generalize to future trading. By contrast, our self-supervised SRL approach pays closer attention to the quality and comprehensiveness of the news, which could help with out-of-distribution generalization on the realistic trading.

\subsection{Semantic Role Labeling and Self-Supervised Learning Approach}
Semantic role labeling (SRL) aims to disclose the predicate-argument structure of a given sentence, which could provide a clear overlay that uncovers the underlying semantics of text \cite{conia2021invero}. 
However, previous stock movement prediction methods \cite{xu-cohen-2018-stock,hu2018listening,ding2015deep} adopted the word or sentence level representation to predict the stock movement. Due to the lack of abstract information of the news, these approaches can overfit the training data and fail to distinguish the key features of news. To deal with this problem, we used the SRL's characteristics for extracting a clear overlay that uncovers the underlying semantics of news. \par

Recently, self-supervised learning has become a very popular technique in the training stage of NLP, which generates labels without any human intervention and learns common language representations. Some researches  \cite{im-etal-2021-self,zheng2021self} have proven that self-supervised learning strengthens the generalization ability for models as it improves the performance in many tasks.

\section{Dateset Creation}

\begin{table}[h]
\centering
\caption{The comparison between Astock and other existing widely-used stock prediction dataset.}
\resizebox{1\linewidth}{!}{%
\begin{tabular}{|c|c|c|c|c|}
\hline
Dataset                    & Num of Stock & Text Source & Price-level                                                    & Stock Factors \\ \hline
\vtop{\hbox{\strut DMFT's dataset}\hbox{\strut \cite{zhang2017stock} {\tiny EMNLP 17'} }}           & 50           &     \XSolidBrush          & Daily                                                          & \XSolidBrush                \\ \hline
\vtop{\hbox{\strut StockNet's dataset }\hbox{\strut \cite{xu-cohen-2018-stock} {\tiny ACL 18'}  }}     & 88           & Twitter     &   \XSolidBrush                                                               &        \XSolidBrush         \\ \hline
\vtop{\hbox{\strut Dingxia's dataset }\hbox{\strut  \cite{ding-etal-2014-using} {\tiny EMNLP 14'} }}      & 500          & News        & Daily                                                          &   \XSolidBrush              \\ \hline
\vtop{\hbox{\strut Trade the event 's dataset  }\hbox{\strut   \cite{zhou-etal-2021-trade} {\tiny ACL 21'} } } &        \XSolidBrush        & News        &     \XSolidBrush                                                             &  \XSolidBrush               \\ \hline
Ours                       & 3680         & Stock News  & \makecell{Minute-level when news published \\  Daily-level for all the stocks} &      \CheckmarkBold         \\ \hline
\end{tabular}
}
\label{comparisondata}
\end{table}
The stock prediction task aims to explore a realistic method to predict the stock movement with comprehensive and reasonable information in the China stock market. 
To this end, it is important to have minute-level price information in the dataset and we are motivated to collect one.


\subsection{Standard of news and stock factors collection}
\label{factorrr}
There are two main components in our dataset: News and stock factors for the China stock market. In terms of news data, there are 40,963 pieces of listed company news, including company announcements and company-related news from July 2018 to November 2021. The news data are split into two parts: the In-distribution split and the out-of-distribution split. The in-distribution split is from July 2018 to December 2020 for training and testing where the training set occupies by 80\%, and the validation set and test set occupy 10\% respectively. The out-of-distribution split is selected from January 2021 to November 2021, which is used for OOD generalization testing. Every piece of news includes its published time and a corresponding news summary. Factor investing is an investment approach that involves targeting quantifiable firm characteristics or factors that can explain the differences in stock returns. 
Factor-based strategies may help investors
meet particular investment objectives—such as
potentially improving returns or reducing risk
over the long term. Our Astock dataset covers the 24 stock factors on each stock of the China A-shares including  Dividend yield, Total share, Circulated share, Free Float share, Market Capitalization, Price-earning ratio, PE for Trailing Twelve Months, Price/book value ratio, Price-to-sales Ratio, Price to Sales ratio, Circulate Market Capitalization, Open price, High price,Low price, Close price , Previous close price, Price change, Percentage of change, Volume, Amount, Turn over rate, Turn over rate for circulated Market Capitalization, Volume ratio. Furthermore, We compare Astock with several widely used stock prediction datasets in Table \ref{comparisondata}. The value is reflected in the following aspects: (1) Astock provides financial news for each specific stock over the entire China A-shares market. (2) Astock provides various stock factors for each stock. (3) Astock provides minute-level historical prices for the news.
\label{traindate}

\subsection{Task Formulation}
 We divide the automated trading system into two tasks: stock movement classification and simulated trading.

\subsubsection{Text-based stock movement classification}

The goal of the stock movement classification task is to classify the effects of the input information. We measure the impact of each piece of company news by the stock return rate. 
 In this paper, the news is annotated by the stock return rate $r$ , and three cases are considered in our annotation: outperforming, neutral, and underperforming as shown in Equation \ref{labels}. We further model the stock movement by classifying it into three categories. The ground truth for those categories can be derived from $r$. Specifically, we follow the following rules to categorize the data into three classes after ranking all the news by $r$, which aims to find the most strong signal of the stock movement, and to reduce the disturbance of noises comparing to dividing the data evenly. After the domain experts gave us the advice and the experiments with different thresholds was conducted, we set 20\% as the threshold where the tunable parameters a, b, c, and d are 20, 40, 60, and 20, respectively.
 
\begin{equation}
    label= 
\begin{cases}
      outperforming,      &\text{if } r \; ranked \; top\ a\%\   \\
    neutral,      &\text{if } r  \; ranked\; top\ b\%-c\%\ \\
    underperforming,      &\text{if } r  \;ranked\;  bottom\ d\%\ 
\end{cases}
\label{labels}
\end{equation}
where $r$ is the return rate of the news. We randomly select 80\% of the in-distribution dataset as the training set, and the other 20\% is split evenly into validation and test sets. 


\subsubsection{Simulated Trading}
Stock movement prediction accuracy may not necessarily translate to a profitability of an auto-trading system. To further investigate how the stock prediction can benefit for the actual trading practice, we employ a practical trading strategy based on the stock movement prediction results and evaluate various metrics for the trading actions. The trading strategy details can be found at our github page.


\section{Methodology}
This section describes the technical contribution of this work: a novel system for stock movement prediction. Our system consists of two major components: semantic role labeling pooling method and a self-supervised learning based on SRLP, we will elaborate on those two parts. 

\begin{figure*}[t]
\begin{center}
\resizebox{0.7\textwidth}{!}{%
\includegraphics[width=0.8\linewidth]{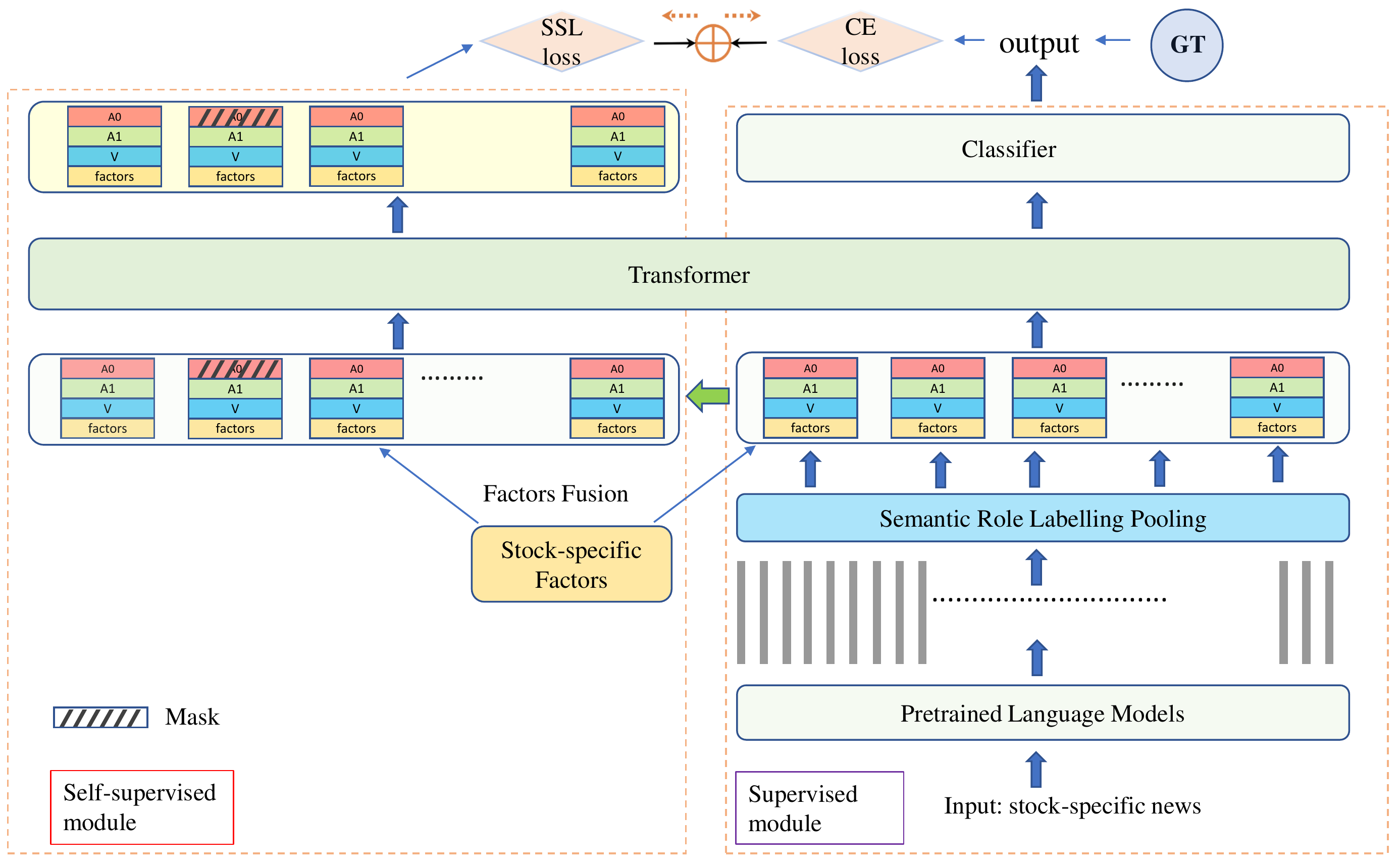}}
\end{center}
\caption{\label{zero} Overall framework of our approach, including a domain adapted pre-trained model (RoBERTa WWM Ext), Semantic Roles Pooling, transformer layer, self-supervised module (left part), and the supervised module (right part). The green arrow represents a duplicate for the SRLP. The final result is generated from the stock movement classifier, and the total loss is obtained from the self-supervised SRLP part and supervised stock movement classification part.}
\label{yyds}
\end{figure*}

\subsection{Semantic Role Labelling Pooling}
\begin{figure}[H]
\begin{center}  
   \includegraphics[width=\linewidth]{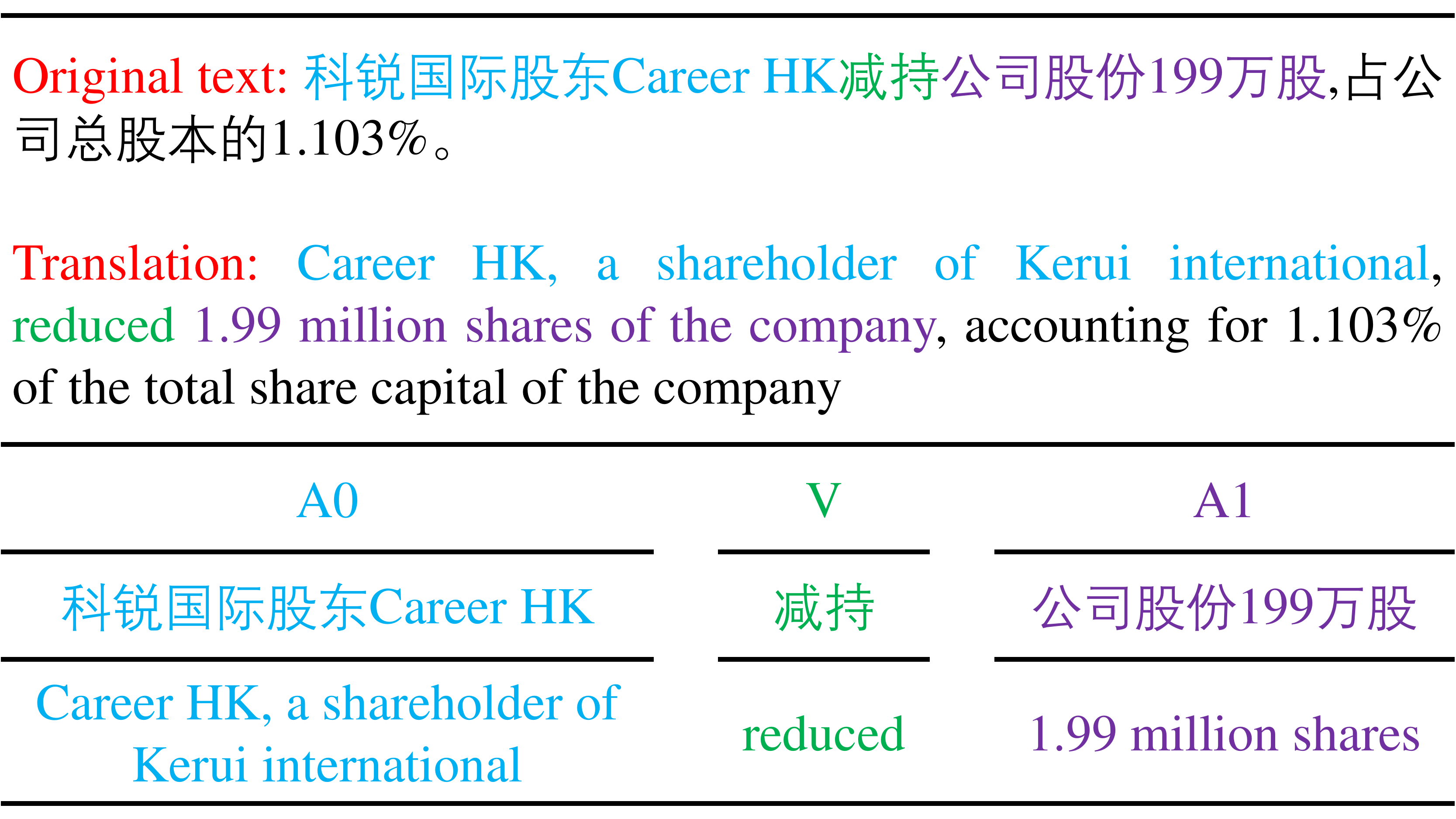}
\end{center}
\caption{A Semantic role labeling example for a piece of news.}
\label{srlexample}
\end{figure}

In this work, we propose to leverage the off-the-shelf semantic role labeling, i.e., Propbank \cite{kingsbury2003propbank}, to pool the output embeddings of a pre-trained language model to construct an alternative representation. The rationale is that the semantic roles in Propbank, i.e., verb (V), proto-agent (A0), and proto-patient (A1), are general-purposed and are also strongly associated with the event arguments. We show an example for semantic role labeling for financial news in Figure \ref{srlexample}.

More specifically, we first use the Language Technology Platform (LTP) \cite{LTP} to automatically mark the semantic roles from the sentences of an entire piece of news and then select V, A0, and A1 to represent the roles for each sentence. Secondly, we process each sentence with a pretrained language model to obtain a sequence of output embeddings
$\{\mathbf{s}_1,\mathbf{s}_2,\cdots,\mathbf{s}_n\}$. We use $\mathcal{V}$, $\mathcal{A}_0$ and $\mathcal{A}_1$ to denote the indices of tokens corresponding to the V, A0, A1 components. At last, we perform pooling for embeddings with their indices falling into $\mathcal{V}$, $\mathcal{A}_0$ and $\mathcal{A}_1$. We call this scheme Semantic Role Labelling Pooling SRLP in short. Taking A0 as an example, the SRLP feature for A0 is 
\begin{align}
\mathbf{e}_{A0} = \frac{1}{|\mathcal{A}_0|}\sum_{i\in \mathcal{A}_0 }\mathbf{s}_i
\end{align}

For a sentence with $N$ sets of V, A0 and A1, we concatenate $\mathbf{e}_{A0}$,$\mathbf{e}_{A1}$,$\mathbf{e}_{V}$ of each sentence and the financial factor $\mathbf{F}$ of the stock-of-interest into a data matrix: 
\begin{equation}
\begin{aligned}
\mathbf{E} = \begin{bmatrix}
                \mathbf{e}_V^1  & ...& \mathbf{e}_V^t& ...& \mathbf{e}_V^N\\
                \mathbf{e}_{A0}^1  & ...& \mathbf{e}_{A0}^t & ...& \mathbf{e}_{A0}^N \\
                \mathbf{e}_{A1}^1  & ...& \mathbf{e}_{A1}^t& ...& \mathbf{e}_{A1}^N \\
                \mathbf{F}  & ...& \mathbf{F} & ...& \mathbf{F},
            \end{bmatrix}
\end{aligned}
\label{eq.srl}
\end{equation}
where $\mathbf{E}$ is output of the above process. Each column of $\mathbf{E}$, denoted as $\mathbf{e}_j$, is the concatenation of $\mathbf{e}^j_V,\mathbf{e}^j_{A0},\mathbf{e}^j_{A1}$ and $\mathbf{F}$. $\mathbf{E}$ is then processed by a Transformer encoder in the same way as the standard text classification to generate the stock movement prediction.


\subsection{Self-Supervised Learning based on SRLP}
Besides standard supervised training loss for stock movement classification, in this work, we further propose to use a self-supervised training task as an auxiliary task to train the network. For stock prediction, good generalization is highly desirable since the training data is usually sampled from a period different from the test period.
\begin{table*}[t]
\centering
\caption{The stock movement classification performance(\%) of in-distribution evaluation on our scheme and others demonstrates the effectiveness of our self-supervised SRL method.  \CheckmarkBold \  indicates that the model adopted this Semantic role's pooling information.  \textbf{-} indicates that the method does not adopt this semantic role's pooling. \XSolidBrush \ indicates that semantic role's pooling is masked.}
\label{mainresult}
\resizebox{0.80\textwidth}{!}{%
\begin{tabular}{c|c|ccc|c|cl|c|c}
\hline
\multirow{2}{*}{Model}          & \multirow{2}{*}{Resource} & \multicolumn{3}{c|}{ Semantic Role}                                                & \multirow{2}{*}{Accuracy} & \multicolumn{2}{c|}{\multirow{2}{*}{F1 Score}} & \multirow{2}{*}{Recall} & \multirow{2}{*}{Precision} \\ \cline{3-5}
                                &                           & \multicolumn{1}{c|}{A0}         & \multicolumn{1}{c|}{V}          & A1         &                           & \multicolumn{2}{c|}{}            &                         &                            \\
                                \hline
                                
\makecell{StockNet \cite{xu-cohen-2018-stock} {\tiny ACL 18'}}              & News                      & \multicolumn{1}{c|}{\textbf{-}}          & \multicolumn{1}{c|}{\textbf{-}}          & \textbf{-}         &46.72                     & \multicolumn{2}{c|}{44.44}                    &46.68                  &47.65                     \\ \hline
\makecell{HAN Stock \cite{hu2018listening} {\tiny ICWSDM 18'}}              & News                      & \multicolumn{1}{c|}{\textbf{-}}          & \multicolumn{1}{c|}{\textbf{-}}          & \textbf{-}          &     57.35            & \multicolumn{2}{c|}{56.61}                    &        57.20          &      58.41               \\ \hline

\makecell{Bert Chinese \cite{devlin-etal-2019-bert}{\tiny NAACL 19'}}         & News                      & \multicolumn{1}{c|}{\textbf{-}}          & \multicolumn{1}{c|}{\textbf{-}}          & \textbf{-}         & 59.11                    & \multicolumn{2}{c|}{58.99}                    & 59.20                  & 59.07                     \\ \hline

\makecell{ERNIE-SKEP \cite{tian-etal-2020-skep} {\tiny ACL 20'}}           & News                      & \multicolumn{1}{c|}{\textbf{-}}          & \multicolumn{1}{c|}{\textbf{-}}          & \textbf{-}         & 60.66                    & \multicolumn{2}{c|}{60.66}                    & 60.59                  & 61.85                     \\ \hline

\makecell{XLNET Chinese \cite{cui-etal-2020-revisiting}{\tiny EMNLP 20'}}                & News                      & \multicolumn{1}{c|}{\textbf{-}}          & \multicolumn{1}{c|}{\textbf{-}}          & \textbf{-}          & 61.14                    & \multicolumn{2}{c|}{61.19}                    & 61.09                  & 61.60                     \\ \hline

\makecell{RoBERTa WWM Ext \cite{cui-etal-2020-revisiting}{\tiny EMNLP 20'}}              & News                      & \multicolumn{1}{c|}{\textbf{-}}          & \multicolumn{1}{c|}{\textbf{-}}          & \textbf{-}          & 61.34                    & \multicolumn{2}{c|}{61.48}                    & 61.32                  & 61.97                     \\
\makecell{ }      & News + Factors                       & \multicolumn{1}{c|}{\textbf{-}}          & \multicolumn{1}{c|}{\textbf{-}}          & \textbf{-}          & 62.49                    & \multicolumn{2}{c|}{62.54}                    & 62.51                  & 62.59                    \\ \hline
\makecell{Our SRLP}                  & News                      & \multicolumn{1}{c|}{\CheckmarkBold} & \multicolumn{1}{c|}{\CheckmarkBold} & \CheckmarkBold & 61.76                    & \multicolumn{2}{c|}{61.69}                    & 61.62                  & 61.87                     \\
\makecell{}        & News + Factors            & \multicolumn{1}{c|}{\CheckmarkBold} & \multicolumn{1}{c|}{\CheckmarkBold} & \CheckmarkBold & 64.79                    & \multicolumn{2}{c|}{64.85}                    & 64.79                  & 65.26                     \\ \hline
\makecell{Our Self-supervised SRLP}  & News                      & \multicolumn{1}{c|}{\XSolidBrush}           & \multicolumn{1}{c|}{\CheckmarkBold} &     \XSolidBrush      & 61.07                   & \multicolumn{2}{c|}{61.11}                    & 61.11                 & 61.11                     \\
                                & News                      & \multicolumn{1}{c|}{\XSolidBrush} & \multicolumn{1}{c|}{\CheckmarkBold}           &     \CheckmarkBold       & 62.36                    & \multicolumn{2}{c|}{62.32}                    & 62.43                  & 62.64                     \\
                                & News                      & \multicolumn{1}{c|}{\CheckmarkBold}           & \multicolumn{1}{c|}{\CheckmarkBold}           & \XSolidBrush & 62.42                    & \multicolumn{2}{c|}{62.46}                    & 62.44                  & 62.62                     \\
                                & News                      & \multicolumn{1}{c|}{\XSolidBrush} & \multicolumn{1}{c|}{\XSolidBrush} &     \CheckmarkBold       & 62.15                    & \multicolumn{2}{c|}{62.15}                    & 62.15                  & 62.59                     \\
                                & News                      & \multicolumn{1}{c|}{\CheckmarkBold}           & \multicolumn{1}{c|}{\XSolidBrush} & \XSolidBrush & 61.34                    & \multicolumn{2}{c|}{61.23}                    & 61.46                  & 61.30                     \\
                                & News                      & \multicolumn{1}{c|}{\CheckmarkBold} & \multicolumn{1}{c|}{\XSolidBrush}           & \CheckmarkBold & 62.97                   & \multicolumn{2}{c|}{63.05}                    & 62.93                  & 63.47                    \\ \hline
\makecell{Our Self-supervised SRLP} & News + Factors            & \multicolumn{1}{c|}{\XSolidBrush}           & \multicolumn{1}{c|}{\CheckmarkBold} &    \XSolidBrush  & 64.59        & \multicolumn{2}{c|}{64.62}           &  64.63   & 64.65           \\                     
with Factors                       & News + Factors            & \multicolumn{1}{c|}{\XSolidBrush} & \multicolumn{1}{c|}{\CheckmarkBold}           &    \CheckmarkBold               & 66.82          & \multicolumn{2}{c|}{66.81}           &  66.90   & 66.82           \\    
                                & News + Factors            & \multicolumn{1}{c|}{\CheckmarkBold}           & \multicolumn{1}{c|}{\CheckmarkBold}           & \XSolidBrush & 65.54                    & \multicolumn{2}{c|}{65.53}                    & 65.62                  & 65.50                     \\
                                & News + Factors            & \multicolumn{1}{c|}{\XSolidBrush} & \multicolumn{1}{c|}{\XSolidBrush} &    \CheckmarkBold        & 65.34                    & \multicolumn{2}{c|}{65.21}                    & 65.43                  & 65.43                     \\
                                & News + Factors            & \multicolumn{1}{c|}{\CheckmarkBold}           & \multicolumn{1}{c|}{\XSolidBrush} & \XSolidBrush & 65.27                    & \multicolumn{2}{c|}{65.35}                    & 65.24                  & 65.77                     \\
                                & News + Factors            & \multicolumn{1}{c|}{\CheckmarkBold} & \multicolumn{1}{c|}{\XSolidBrush}           & \CheckmarkBold & \textbf{66.89}                   & \multicolumn{2}{c|}{\textbf{66.92}}                    & \textbf{66.95}                 &  \textbf{66.92}   
                                \\ \hline
\end{tabular}%
}
\label{mainresult}
\end{table*}
A significant problem in practice is to ensure that our model generalizes to scenarios different from the training set. We further create a self-supervised learning method on top of the SRLP. Recent studies \cite{mohseni2020self,hendrycks2019selfsupervised} have shown that incorporating a self-supervised learning task along with the supervised training task could lead to better generalization. As shown in Figure \ref{yyds}, the self-supervised task is defined as predicting the position of one randomly masked SRL role from all the roles of SRL in a piece of news. Intuitively, the self-supervised learning task should be designed to encourage the favorable properties of features. In this work, we propose to randomly mask one pooled embedding, i.e., $\mathbf{e}_{V}^j$,  $\mathbf{e}_{A0}^j$ or $\mathbf{e}_{A1}^j$, from a randomly selected sentence, and then ask the network to identify the masked embedding from a pool of candidate embeddings. Such a cloze-style task encourages the network to perform reasoning over other unmasked cues to work out the missing item. We hypothesize that such a reasoning capability is beneficial for understanding the financial news and thus helps stock movement prediction. 

Formally, we randomly select a $\mathbf{e}_j$ from $\mathbf{E}$ and then select one element from $\mathbf{e}_j$ = \{  $\mathbf{e}_{V}^j$,  $\mathbf{e}_{A0}^j$,  $\mathbf{e}_{A1}^j$\}, after that we replace the selected element with an all-zero vector, indicating a "mask" operation. 
Taking masked V at the t-th sentence as an example, we denote the $\mathbf{E}$ after this mask operation as $\mathbf{E}'$.
\[
\begin{aligned}
\mathbf{E}^{\prime} = \begin{bmatrix}
                \mathbf{e}_V^1  & ...& \mathbf{M}& ...& \mathbf{e}_V^N\\
                \mathbf{e}_{A0}^1  & ...& \mathbf{e}_{A0}^t & ...& \mathbf{e}_{A0}^N \\
                \mathbf{e}_{A1}^1  & ...& \mathbf{e}_{A1}^t& ...& \mathbf{e}_{A1}^N \\
                \mathbf{F}  & ...& \mathbf{F} & ...& \mathbf{F} 
            \end{bmatrix}
\end{aligned}
\label{eq.srl}
\]
Then we feed $\mathbf{E}'$ into the transformer to obtain a query vector sequence $\mathbf{q} \in \mathbb{R}^d$
\begin{align}
    \textbf{q} = Transformer(\textbf{E}^{\prime})[:,t] \nonumber
    \label{eq.qk}
\end{align}
where $[:,t]$ means extract the t-th column of the vector sequences calculated by the transformer. The unmasked SRLP-V features (or SRLP-A0, SRLP-A1 features, depending on which type of SRLP feature is chosen) is also send to an encoder to calculate candidate key vectors:
Formally, $\mathbf{K}$ is defined as:
\begin{align}
\mathbf{K} =  [f_V(\mathbf{e}_V^1), \cdots, f_v(\mathbf{e}_V^t), \cdots, f_V(\mathbf{e}_V^N)] \in \mathbf{R}^{d\times N}\nonumber
\end{align} where $f_V$ is an encoder specified for encoding V-type SRLP feature. Then the query vector is compared against each column vector in $\mathbf{K}$ and is expected to have the highest matching score at the $t$-th location. This process could be implemented via matrix multiplication and the softmax operation:

\begin{equation}
    P_{SSL} = \operatorname{Softmax}(\mathbf{q}\mathbf{K})
    \label{eq.qk}
\end{equation}
and we hope the highest probability entry in Eq. 
\ref{eq.qk} is at the $t$-th dimension. This requirement could be enforced via the cross-entropy loss. Finally, the training loss for the models is 
\begin{equation}
    \mathcal{L} = \alpha \mathcal{L}_{CLS}+ (1-\alpha)  \mathcal{L}_{SSL}
    \label{eq.loss}
\end{equation}
where $\mathcal{L}_{CLS}$ is the cross-entropy loss for the text classification and $\mathcal{L}_{SSL}$ is the cross-entropy loss on the self-supervised learning prediction $P_{SSL}$. $\alpha$ here is a trade-off parameter.

\section{Experiments}
In this section, we conduct experiments to evaluate the performance of the proposed model. We conduct experiments on two different splits of our dataset for each model: In-distribution split and out-of-distribution split. We also feed the prediction result of our method to the proposed trading strategy to analyze the profitability through backtesting on real-world stock data.

\subsection{Evaluation metrics}
We evaluated the stock movement prediction and simulated trading performance. For the Stock movement prediction, we applied the \textbf{Accuracy, F1 Score, Recall and Precision} as evaluation metrics. For simulated trading, we applied the \textbf{Annualized Rate of Return, Maximum Drawdown and Sharpe Ration} as evaluation metrics based on our simulated trading strategy.

\subsection{Compared Methods}
We re-implement the current state-of-art stock movement prediction models as baselines, including StockNet \cite{xu-cohen-2018-stock}, HAN Stock \cite{hu2018listening}. \textbf{StockNet} \cite{xu-cohen-2018-stock} is a stock temporally-dependent movement prediction model which also uses Twitter data and price information to predict two classes for stock movement. Hybrid Attention Networks \textbf{(HAN) }Stock \cite{hu2018listening} is a stock trend prediction model based on a sequence of recent related news to predict three classes for stock movement task, which are the same as ours. We also construct baselines by formulating the stock movement prediction problem as text classification and use four strong pre-trained Chinese language models as backbones such as \textbf{XLNet-base-Chinese}\cite{cui-etal-2020-revisiting}, Sentiment Knowledge Enhanced pre-trained language model(\textbf{SKEP})\cite{tian-etal-2020-skep}, \textbf{Bert Chinese}\cite{devlin-etal-2019-bert} and \textbf{RoBERTa WWM Ext}\cite{cui-etal-2020-revisiting}. For the above four pre-trained language models, we extract sentence embedding from the [CLS] token and attach a three-way classifier to predict the stock movements. In addition, we also compare the CSI300 index, XIN9 index\footnote{Equivalent to the Standard and Poor's 500 (S\&P 500) or the Dow Jones Industrial Average (DJIA) in the US stock market}
against the proposed method when analyzing the profitability of the proposed system. For our methods, we used RoBERTa WWM Ext as our backbone PLM since the remarkable performance.

\subsection{Stock Movement Evaluation}
We first compare different methods on the task of stock movement prediction. We conduct experiments on two different splits of our dataset: In-distribution split and out-of-distribution split. In the in-distribution split, both training and testing data are sampled from the same period while the out-of-distribution split uses data from different periods to construct the training and testing data. \par

\noindent \textbf{In-distribution evaluation} 

 The results are shown in Table  \ref{mainresult}. From the results, we made the following two observations:

1. If only text information is used, the proposed SRLP approach achieves the state-of-the-art performance. Interestingly, we find that SRLP achieves superior performance when further combines the stock factors. It outperforms RoBERTa WWM Ext (News+Factors) by more than 2\%. We postulate that this is because the compact representation in SRLP make incorporation of stock factors easier. Note that the proposed way of incorporating stock factors (see Section \ref{factorrr}) does not only introduce extra modalities for the stock movement prediction but also could make the text analysis module adaptive to the stock factors. This could be useful to model the scenario like the effect of a similar event could result in a different impact on the stock movement for a different type of company.\par
2. The proposed self-supervised SRLP can further boost the performance of SRLP. In the best setting of self-supervised SRLP, i.e., with V being masked, self-supervised SRLP achieves more than 1\% improvement over SRLP.  The improvement is even larger when the stock factors are provided, showing more than 2\% improvement over SRLP (News+Factors), achieving 66.89\% prediction accuracy. This validates the effectiveness of the proposed self-supervised learning approach. Interestingly, we observe that masking A0 and A1 usually will not bring improvement in contrast to the case of masking V. Note that V encodes the type of an event, and the argument is encoded by A0 or A1. It seems that predicting the type of events is a more effective self-supervised learning task than working on the argument.

\noindent \textbf{Out-of-distribution evaluation}

In the experiments above, the training data and testing data are sampled from the same period. Thus the distributions of training data and testing data are similar. For real-world applications, the stock movement prediction model is applied to future data unseen at the training time. Hence, it is critical to evaluate the model in such an out-of-distribution setting. 
\begin{table}[H]
\centering
\caption{The comparison (\%) of the out-of-distribution evaluation on stock movement classification with StockNet, RoBERTa-WWM Ext, HAN Stock method and our method from 1/1/2021 to 12/11/2021.}
\resizebox{\linewidth}{!}{%
\begin{tabular}{l|c|c|c|c}
\hline
\multicolumn{1}{c|}{Model}    &\makecell{Accuracy}   & \makecell{F1 Score} & \makecell{Recall} & \makecell{Precision} 
\\
\hline
StockNet\cite{xu-cohen-2018-stock}   & 44.35  &42.52   & 45.42   & 45.82  \\
HAN Stock\cite{hu2018listening}   & 53.41  & 53.33   & 53.69   & 54.53  \\
RoBERTa WWM Ext\cite{cui-etal-2020-revisiting} & 60.15    & 60.08      & 59.89 & 60.78 
\\

Self-supervised SRLP(V masked)+Factors & \textbf{64.09} & \textbf{63.95}                             & \textbf{63.90}           & \textbf{64.43}                    \\ 
\hline
\end{tabular}
 }
\label{ood2}
\end{table}
To this end, we construct a new training/testing split by using the data from July 2018 to December 2020 as training data and the data from January 2021 to November 2021 for the testing data. We first conduct an evaluation on the stock movement prediction task, and the results are shown in Table  \ref{ood2}. From the results, we can see that the proposed method is still comparably competitive over other baselines. 




\subsection{Profitability Evaluation in Real-world}

\begin{table}[H]
\centering
\caption{The comparison of profitability test on Maximum Drawdown(\%), Annualized Rate of Return(\%), and Sharpe Ratio Rate(\%) with strong baselines, XIN9, CSI300 and our proposed method from 1/1/2021 to 12/11/2021.}
\resizebox{\linewidth}{!}{
\begin{tabular}{l|c|c|c}
\hline
\multicolumn{1}{c|}{Model}      & \makecell{Maximum \\ Drawdown}$\downarrow$ & \makecell{Annualized Rate \\ of Return}$\uparrow$  & \makecell{Sharpe \\ Ratio}$\uparrow$ \\ \hline
XIN9                            & -15.85                               & -15.38                                   & -32.01                           \\
CSI300                         & -14.40                      & -9.34                                & -32.99                          \\
StockNet\cite{xu-cohen-2018-stock}                        & -7.40                                        & -22.42                   & -177.65                          \\
HAN Stock\cite{hu2018listening}                        & -7.38                      & -13.50                                  & -55.84                          \\
RoBERTa WWM Ext\cite{cui-etal-2020-revisiting}                     & -3.83                               & 1.35                                    & -16.31                          \\

Self-supervised SRLP(V masked) with Factors     & \textbf{-3.60}                               & \textbf{13.85}                                  &\textbf{40.93}                      \\
\hline 
\end{tabular}
  }
\label{returnperformance}
\end{table}

\begin{figure}[]
\begin{center}  
  \includegraphics[width=\linewidth]{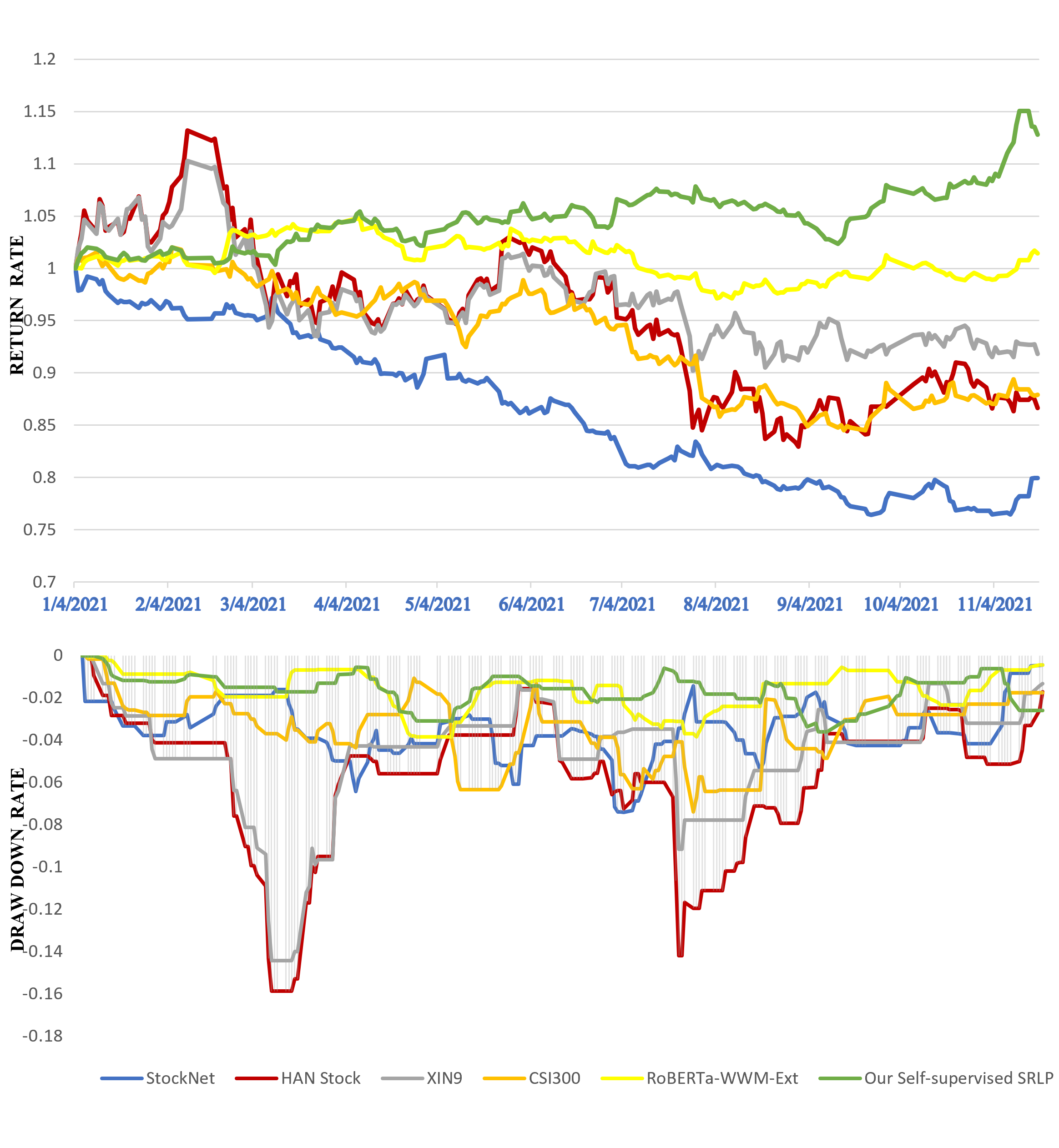}
\end{center}
\caption{The comparison for the real trading performance on Return Rate, Draw Down Rate with  CSI300 index, XIN9, Roberta WWM Ext, HAN Stock, StockNet and our proposed method from 1/1/2021 to 12/11/2021}
\label{return}
\end{figure}

In this section, we discuss the possible profitability of the proposed strategy in real-world trading.  We use our trading strategy to conduct trading simulation (backtesting) on stock data from January 2021 to November 2021 using the stock movement prediction result of our model trained from July 2018 to December 2020 as mentioned in Section \ref{traindate}.  In Table \ref{returnperformance}, we show that our self-supervised SRLP model achieves a remarkable annualized rate of return of 13.85\% , which surpasses the previous baselines and market index XIN9 and CSI300.  The resulting baseline HAN Stock \cite{hu2018listening} and StockNet \cite{xu-cohen-2018-stock} achieve an annualized rate of return of -13.5\% and -22.42\% respectively, and the market XIN9 index and CSI300 were overall declining in 2021, which obtains -15.38 \% and -9.34\% respectively.  In addition, our self-supervised learning method also obtains the lowest Maximum Drawdown of -3.6\% and the highest Sharpe Ratio of 40.93\% , which significantly 
outperforms the previous methods and indicates that our self-supervised could successfully achieve higher expected returns while remaining relatively less risky as shown in Figure \ref{return}. 


\section{Conclusion}


In this paper, we study the problem of NLP-based stock prediction and build a platform with a new dataset, AStock, featured by: (1) Large number of stocks, stock-relevant news. (2) Availability of various financial factors. (3) Financial-relevent metrics for Evaluation. The Platform is based on two novel techniques. One leverages Propbank-style semantic role labeling results to create compact news representation. Building on top of this representation, the other technique is a customized self-supervised learning training strategy for improving generalization performance. We demonstrate that the proposed method achieves superior performance over other baselines through extensive experiments in both in-distribution and out-of-distribution settings. Also, by feeding our prediction to a practical simulated trading, our method achieves better profitability in backtesting.

\bibliographystyle{named}
\bibliography{custom,anthology}

\end{document}